\documentclass{article}

\usepackage[preprint]{neurips_2024}

\usepackage[utf8]{inputenc} 
\usepackage[T1]{fontenc}    
\usepackage{hyperref}       
\usepackage{url}            
\usepackage{booktabs}       
\usepackage{amsfonts}       
\usepackage{nicefrac}       
\usepackage{microtype}      
\usepackage{xcolor}         
\usepackage{amsmath}
\usepackage{amssymb}
\usepackage{nomencl}
\usepackage{glossaries}
\usepackage{xspace}
\usepackage{multirow}
\usepackage{graphicx}
\usepackage{bbding}
\usepackage{pifont}
\usepackage{algorithm}
\usepackage{subfigure}
\usepackage{enumitem}
\usepackage{multicol}
\usepackage{array}
\usepackage{fp}
\usepackage{makecell}
\usepackage{flushend}
\usepackage{cases}
\usepackage{color}
\usepackage{listings}
\usepackage{mathrsfs}
\usepackage{longtable}
\usepackage{colortbl}
\usepackage{bm}
\usepackage{algorithm}
\usepackage{algorithmic}

\definecolor{codegreen}{rgb}{0,0.3,0.6}
\definecolor{codegray}{rgb}{0.5,0.5,0.5}

\newcommand{\ie}{\emph{i.e.,}\xspace}
\newcommand{\eg}{\emph{e.g.,}\xspace}

\newcommand{\paratitle}[1]{\vspace{1.5ex}\noindent\textbf{#1}}

\newcommand{\ignore}[1]{}

\usepackage{tcolorbox}
\definecolor{darkorange}{RGB}{255, 140, 0}
\definecolor{lightgreen}{RGB}{145, 204, 117}
\definecolor{lightyellow}{RGB}{250, 200, 88}
\definecolor{lightred}{RGB}{238, 102, 102}
\definecolor{lightblue}{RGB}{115, 192, 222}
\newtcolorbox{promptbox}[2][Prompt]{
colback=black!5!white,
arc=5pt, 
boxrule=0.5pt,
fonttitle=\bfseries,
title=#1, 
before upper={\scriptsize}, fontupper=\fontfamily{ptm}\selectfont,
colframe=#2, 
}

\title{Think More, Hallucinate Less: Mitigating Hallucinations via Dual Process of Fast and Slow Thinking}
\author{
    Xiaoxue Cheng$^{1}$\thanks{Equal contribution.}~,
  Junyi Li$^{2*}$, \textbf{Wayne Xin Zhao$^{1}$\thanks{Correspondence to Wayne Xin Zhao.}}~, \textbf{Ji-Rong Wen}$^{1}$
  \\
  $^1$Gaoling School of Artificial Intelligence, Renmin University of China\\
  $^2$Department of Computer Science, National University of Singapore \\
  \texttt{chengxiaoxue@ruc.edu.cn} \quad
  \texttt{junyi\_cs@nus.edu.sg} \quad
  \texttt{batmanfly@gmail.com} \\
}

\begin{document}
Technical Report on Slow Thinking with LLMs: Hallucination Mitigation

\maketitle

\begin{abstract}
Large language models (LLMs) demonstrate exceptional capabilities, yet still face the hallucination issue. Typical text generation approaches adopt an auto-regressive generation without deliberate reasoning, which often results in untrustworthy and factually inaccurate responses.
In this paper, we propose \textbf{HaluSearch}, a novel framework that incorporates tree search-based algorithms (\eg MCTS) to enable an explicit slow thinking generation process for mitigating hallucinations of LLMs during inference.
Specifically, HaluSearch frames text generation as a step-by-step reasoning process, using a self-evaluation reward model to score each generation step and guide the tree search towards the most reliable generation pathway for fully exploiting the internal knowledge of LLMs.
To balance efficiency and quality, we introduce a hierarchical thinking system switch mechanism inspired by the dual process theory in cognitive science, which dynamically alternates between fast and slow thinking modes at both the instance and step levels, adapting to the complexity of questions and reasoning states.
We conduct extensive experiments on both English and Chinese datasets and the results show that our approach significantly outperforms baseline approaches.
\end{abstract}

\section{Introduction}

Large language models (LLMs)~\cite{zhao2023survey} are revolutionizing the landscape of artificial intelligence, showcasing remarkable capabilities in generating human-quality text and tackling diverse language tasks.
Despite these advancements, they often struggle with the issue of hallucination~\cite{ji2023survey, huang2023survey, rawte2023survey, ye2023cognitive, zhang2023siren}, where responses can be untrustworthy or factually inaccurate. This issue significantly impacts the practical applications of LLMs in real-world scenarios. Existing studies~\cite{xu2024hallucination, banerjee2024llms} indicate that due to limitations in training data, model architecture, training method, and other factors, completely eliminating hallucinations is infeasible. Consequently, the development of effective techniques to mitigate hallucinations is critical for improving the reliability and robustness of LLM outputs.

Existing efforts to mitigate hallucinations have targeted different stages of the LLM pipeline, including pre-training~\cite{li2023textbooks}, supervised fine-tuning~\cite{Tian2024fine-tuing, elaraby2023halo, lin2024flame}, and inference~\cite{cove, madaan2024self, Kang2023ever}. In this study, we mainly focus on the hallucination mitigation techniques during the inference stage. Existing approaches can be broadly divided into two categories, \ie retrieval-augmented generation and internal knowledge-based methods.
As the most prominent approach, retrieval-augmented generation (RAG)~\cite{lewis2020retrieval} enhances response accuracy by retrieving documents relevant to the query and incorporating them as additional contextual information~\cite{self-rag}. 
Internal knowledge-based methods, such as step-by-step reasoning~\cite{wei2022chain}, self-verification~\cite{cove}, and self-consistency~\cite{wangself}, rely on prompts to generate intermediate reasoning steps or utilize the model’s consistency by selecting the most coherent response from multiple generated outputs.
Although these studies perform deliberative reasoning to mitigate hallucinations, they operate at the response level and remain constrained by the auto-regressive generation paradigm where intermediate errors can accumulate and potentially lead to incorrect final outputs.


In this paper, we propose \textbf{HaluSearch}, a novel framework that explicitly models response generation as a deliberate thinking process of System 2~\cite{kahneman2011thinking}, incorporating a dynamic system switch mechanism to adaptively alternate between fast and slow thinking modes.
To achieve this goal, we first integrate tree search-based algorithms (\eg MCTS) to formulate text generation as a step-by-step reasoning process, treating each sentence as an individual reasoning step. 
Secondly, inspired by the interaction between System 1 and System 2 in the dual process theory of cognitive science~\cite{wason1974dual}, we propose a dynamic system switch mechanism at both the instance and step levels within the text generation process.
Starting with the input prompt, the dynamic system switch mechanism is employed to determine the appropriate thinking mode for the input question or each reasoning step: fast thinking directly generates a completed response or a single reasoning step, while slow thinking generates multiple intermediate sentences that are evaluated by a reward model. 
Finally, we employ the HaluSearch framework to synthesize preference data to train a reward model for assessing the degree of hallucinations in generated sentences. Given the challenges in training accurate reward model, we explore two approaches: generative reward modeling and critique-based reward modeling, targeting effective self-evaluation to guide the search process.
Compared to previous work, our approach performs step-level reasoning to generate responses rather than relying on response-level refinements, which can achieve more effective and fine-grained hallucination mitigation.


We conduct extensive experiments to evaluate the effectiveness of HaluSearch using Llama3.1-8B-Instruct and Qwen2-7B-Instruct as policy models. The results showcase that our method achieves substantial improvements over previous prompt-based and inference-time intervention baselines across both English and Chinese datasets.
Moreover, we evaluate the self-evaluation reward models trained using generative reward modeling and critique-based reward modeling. Both approaches demonstrate effective hallucination evaluation capabilities. Additionally, we analyze the proposed dynamic system switch mechanism, and the results indicate that it achieves a flexible balance between efficiency and accuracy under different switch thresholds.

Our main contributions can be summarized as:
\begin{itemize}
    \item We propose HaluSearch, which integrates tree search-based algorithms (\eg MCTS) to explicitly implement a slow thinking process during the inference stage of LLMs, fully exploiting their own internal knowledge to mitigate hallucinations in generated text.
    \item We introduce a dynamic system switch mechanism at both instance and step levels, enabling adaptive generation paradigm switch between fast and slow thinking modes during the inference process, thereby improving inference efficiency across various scenarios.
    \item We conduct comprehensive experiments on both English and Chinese datasets, demonstrating that HaluSearch achieves substantial improvements over prompt-based and inference-time intervention methods in mitigating hallucinations and improving response quality. 
\end{itemize}

\section{Related Work}

\subsection{Halluciantion Mitigation}
Studies on hallucination mitigation spans both the training and inference stages of LLMs~\cite{li-etal-2024-dawn}. Many studies~\cite{knowledgeOvershadow, Longpre2024pretrainer, li-etal-2024-dawn} have demonstrated that high-quality training data and techniques such as Reinforcement Learning with Human Feedback (RLHF)~\cite{rlhf} are effective in reducing hallucinations in LLMs.
However, due to the high computational cost and resource requirements of model training, more research has focused on exploring hallucination mitigation methods during the inference stage, which can be broadly divided into two categories.
The first category is retrieval-augmented generation (RAG)~\cite{li-etal-2024-dawn} which reduces hallucinations by retrieving documents relevant to the query and providing them as additional context to the model, relying on external knowledge to improve response accuracy. Another category of methods, including Chain-of-Thought (CoT)~\cite{wei2022chain}, Self-Consistency~\cite{wangself}, and Best-of-N~\cite{lightmanlet}, seeks to mitigate hallucinations by leveraging the internal knowledge of LLMs through prompt-based reasoning or consistency-driven strategies. However, these approaches operate at the response level and are still constrained to the fast and intuitive auto-regressive generation paradigm. 
In contrast, our approach employs MCTS to model the response generation as an explicit step-by-step slow-thinking process, utilizing step-level rewards to explore optimal reasoning paths and produce more reliable responses.

\subsection{System 2 Thinking in LLMs}
System 2 thinking in LLMs emulates the human process of deliberate reasoning to generate high-quality and accurate responses.
Early works primarily implemented System 2 thinking in LLMs by using prompts to guide the generation of intermediate reasoning steps or detailed analysis and reflection, such as Chain-of-Thought (CoT)~\cite{wei2022chain}, Tree-of-Thought (ToT)~\cite{yao2024tree}, and Self-Refine~\cite{madaan2024self}, which have demonstrated notable improvements across various tasks (\eg, question answering and mathematical problem-solving).
Recent approaches focus on incorporating search-augmented reasoning during the decoding process to explicitly implement System 2 thinking.
Many studies \cite{snell2024scaling, wang2024q*, kang2024mindstar, min2024imitate} have demonstrated that scaling inference-time computation serves as an alternative to training for enhancing the performance of LLMs, particularly in complex reasoning scenarios. 
However, the potential of this approach for mitigating hallucinations has not been fully investigated. In this paper, we investigate whether tree search-based slow thinking can effectively leverage accurate internal knowledge from LLMs to mitigate hallucinations and introduce a system switch mechanism to dynamically alternate between System 1 and System 2 thinking modes.
\section{Approach}

\begin{figure*}[tb]
	\centering
	\includegraphics[width=1.00\textwidth]{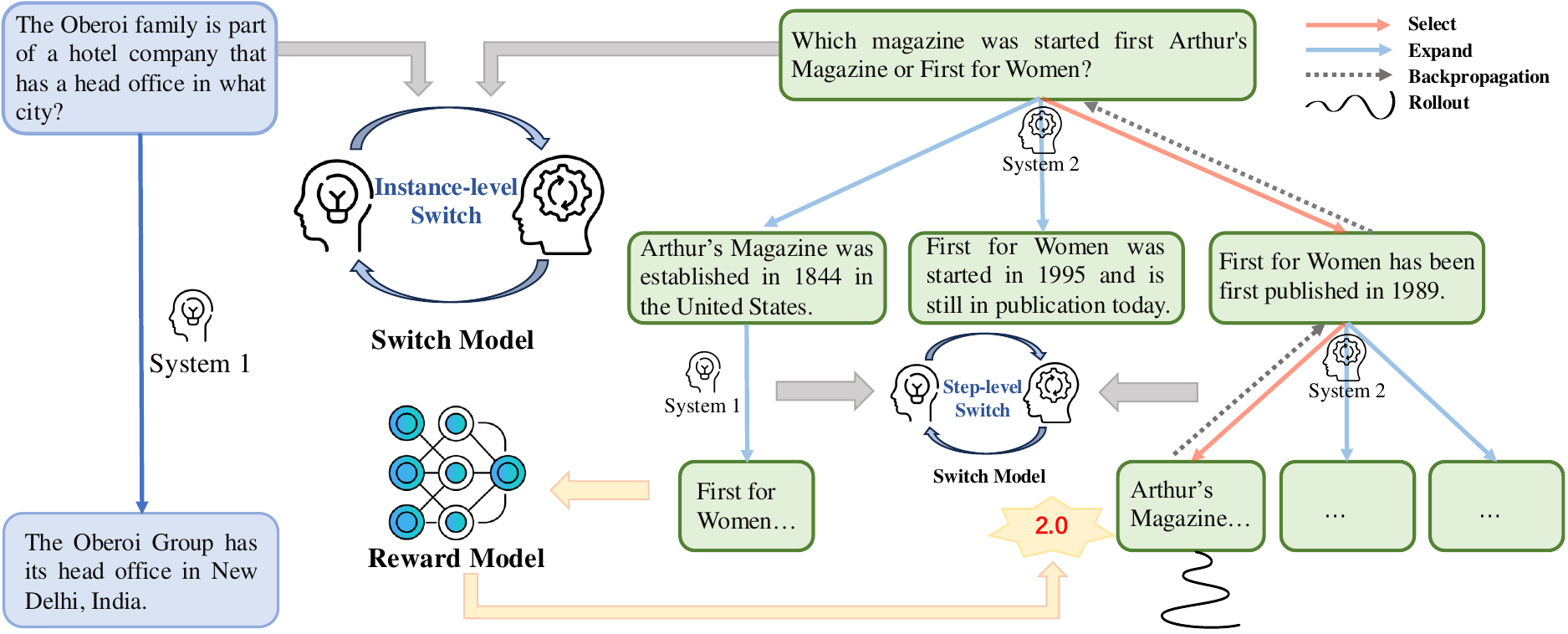}
	\caption{The overview of our proposed HaluSearch approach. The left part demonstrates the process of fast thinking response generation (System 1). The right part illustrates the tree search process with step-level switch between fast thinking and slow thinking generation (System 2).}
	\label{fig: framework}
\end{figure*}

In this paper, we propose \textbf{HaluSearch}, leveraging deliberate reasoning to mitigate LLM hallucinations during the inference stage. Previous work for mitigating hallucinations during inference is mainly limited to the fast thinking paradigm that relies on prompts to instruct LLMs to generate faithful responses or directly calibrates the internal generation mechanism of LLMs~\cite{cove, madaan2024self, li2024inference}.
However, these approaches have not fully exploited the internal knowledge of LLMs to mitigate hallucinations. Some work~\cite{wangself, Orgad2024llms} found that hallucinations often arise from ineffective generation, even when the model possesses knowledge of the underlying facts.
Drawing inspiration from the success of tree search-based algorithms in complex reasoning~\cite{kang2024mindstar, wang2024q*, jiang2024technical}, we propose to integrate Monte Carlo Tree Search (MCTS) with a system switch mechanism between fast and slow thinking modes. Our approach enables an explicit deliberate reasoning process to fully exploit the internal knowledge of LLMs for reducing hallucinations.
The overall architecture of HaluSearch is shown in Figure~\ref{fig: framework}.

\subsection{Problem Formulation}

Our proposed approach aims to mitigate hallucinations in the inference stage. Formally, given an input prompt $x$, the LLM is instructed to generate a response $y = \langle y_1, ..., y_t, ..., y_T \rangle$, where $y_t$ denotes the $t$-th sentence. Specially, we formulate the response generation process of LLMs as step-by-step reasoning, where each sentence represents an intermediate reasoning step. 
In our approach, MCTS aims to construct a search tree $\mathcal{T}$ based on the target LLM. In this tree, a node in the $t$-th tree level is represented as $s_t = \{y_t, N(s_t), V(s_t)\}$, where $y_t$ refers to the generated sentence, $N(s_t)$ denotes the visit count of the node, and $V(s_t)$ represents the value score of the node. The root node $s_0=\{x\}$ only includes the initial input prompt. The target LLM is referred to as the policy model $\pi_\theta$, and the reward model is denoted as $R$.

\subsection{Monte Carlo Tree Search}
\label{sec: MCTS}

In our approach, the MCTS-driven generation process operates as an iterative procedure, where each iteration consists of four key steps: selection, expansion, evaluation, and backpropagation. 
Specifically, the MCTS process begins by initializing the root node of the tree $s_0$ with the input prompt. Subsequently, the four steps are iteratively executed until the predefined termination condition is satisfied. The details of each step are described below.

\paratitle{Selection.} The selection process starts from the root node $s_0$ and selects the leaf node with the highest exploration potential, determined by the UCT (Upper Confidence Bounds applied to Trees)~\cite{uct} score. The UTC score is calculated as follows:
\begin{equation}
    UCT(s_t) = V(s_t) + w \sqrt{\frac{\ln N(p)}{N(s_t)}},
\end{equation}
where $w$ is a hyper-parameter that balances the exploitation (\ie node value $V(s_t)$) and exploration (\ie visit count $N(s_t)$), and $p$ denotes the parent node of $s_t$.

\paratitle{Expansion.} After selecting the node with the highest UCT score, it is expanded by generating multiple child nodes. Based on the historical information, the policy model is employed to generate the next sentence as follows:
\begin{equation}
    y_{t+1} \sim \pi_\theta(\cdot|x, \{y_i\}_{i=1}^t),
\end{equation}
where the previously generated sentences $\{y_i\}_{i=1}^t$ is regarded as the historical context, and the policy model samples and generates $K$ sentences $y_{t+1}$ as a set of child nodes $C(s_{t+1})$. Compared to previous work~\cite{lin2024flame, xie2024improving} mainly focused on a fast thinking generation paradigm, our approach leverages MCTS to explore multiple potential generations for fully exploiting the internal knowledge of LLMs.


\paratitle{Evaluation.} Each expanded node is evaluated to obtain its value $V(s_{t+1})$. Specifically, the policy model performs rollouts to complete the state of each child node $s_{t+1}$ by sampling $m$ completed responses. 
These responses, denoted as $\mathcal{C_\mathrm{r}}(s_{t+1})$, are subsequently evaluated by the reward model (Section~\ref{sec-reward-model}), which assigns a reward score to each response.
The average score $r$ among $m$ completed responses is assigned as the initial value $V(s_{t+1})$ of the corresponding child node $s_{t+1}$.

\paratitle{Backpropagation.} After evaluating the expanded nodes, their values are backpropagated along the traversal path to update the visit counts and value scores of the ancestor nodes $s_j~(0 \leq j \leq t)$. The updates are performed using the following equations:
\begin{align}
    N_{\text{new}}(s_j) &= N_{\text{old}}(s_j) + 1, \\
    V_{\text{new}}(s_j) &= \frac{V_{\text{old}}(s_j)N_{\text{old}}(s_j) + r}{N_{\text{new}}(s_j)},
\end{align}
where $N_{\text{old}}(s_j)$ and $V_{\text{old}}(s_j)$ represent the last visit count and value score of node $s_j$ before backpropagation, respectively, and $r$ is the reward obtained from the evaluation step.

The above four stages are iteratively performed until the policy model reaches the terminal state. We define two kinds of termination conditions for MCTS as follows:
\begin{enumerate}
    \item The maximum iteration $M$ of MCTS is reached.
    \item A terminal node is encountered where the reward satisfies the reward threshold, indicating a lower likelihood of hallucinations.
\end{enumerate}
Once the tree search is completed, the optimal path from the root node to the terminal node is selected using a greedy strategy that prioritizes nodes with the highest value scores and their associated sentences are combined as the final response.

\subsection{Self-Evaluation Reward Model}
\label{sec-reward-model}

In the MCTS-driven generation framework, the reward model plays a crucial role in evaluating the child nodes at each step and guiding the tree search towards more promising directions. While a straightforward approach involves using an advanced LLM (\eg GPT-4) as the reward model, this approach heavily depends on closed-source LLMs. To enable self-evaluation and reduce this dependency, we train the reward model on the same foundation model as the policy model, exploring two reward modeling approaches: generative reward modeling and critique-based reward modeling.

\paratitle{Training Data.} We collect existing question-answering datasets on hallucination evaluation and sample a subset of questions to construct the training data for the reward model. To obtain diverse reward data, we adopt the MCTS generation framework described in Section~\ref{sec: MCTS} to construct the response-score pairs. We explore two reward modeling methods: generative reward modeling and critique-based reward modeling.
For generative reward data, we first collect all complete responses through rollouts and then employ GPT-4 to give a reward score to these completed responses, following the Likert scale~\cite{likert1932technique} ranging from 1 to 5. Each score level is explained with explicit criteria, where higher scores correspond to a greater degree of hallucination. To enhance scoring reliability, the ground truth answer for each question is included in the prompt as a reference for evaluation. 
For critique-based reward data, similar to the aforementioned process, GPT-4 first generates a detailed critique of the response about its correctness and then gives a reward score based on the critique. This format allows the reward model to incorporate critiques as the rationale of its scoring process, facilitating a more accurate evaluation of responses.
To create a balanced training dataset for the reward model, these collected responses are deduplicated based on similarity, and the sample distribution is adjusted to ensure uniform representation across score levels. Detailed scoring guidelines and instructions are shown below.

\begin{promptbox}[Prompt Template for Reward Model]{codegray}
\texttt{Please rate the likelihood of hallucinations (wrong, irrelevant, unfounded, or contradictory content) appearing in the continuation of the current answer fragment. There are five levels of hallucination probability:}\\
\texttt{1 - No hallucination risk: Future content will be entirely accurate, relevant, and well-supported.}\\
\texttt{2 - Low hallucination risk: Future content is likely to be accurate and relevant, with minor uncertainties possible.}\\
\texttt{3 - Moderate hallucination risk: Some hallucinations, such as minor inaccuracies or unclear relevance, may appear, but the content will still be mostly reliable.}\\
\texttt{4 - High hallucination risk: Future content will likely contain noticeable hallucinations, such as errors, irrelevant information, or contradictions, reducing reliability.}\\
\texttt{5 - Very high hallucination risk: Future content is highly likely to include significant hallucinations, such as major errors, contradictions, or fabricated information, making it highly unreliable.}\\
\texttt{Please output a score from 1 to 5. The higher the score, the higher the probability of hallucinations. Only output the score without any further explanation. (Output the score after your analyses.) Do not judge a reply as hallucinated just because it is incomplete. We provide the correct answer as a reference.}\\
\\
\texttt{Question: }\\
\texttt{Correct Answer: \textit{(Only provided when generating reward data.)}}\\
\texttt{Generated Answer:}\\
\texttt{Score:\\}
\end{promptbox}

\paratitle{Training Method.} 
We adopt two kinds of reward modeling approaches, \ie generative and critique-based. Leveraging the generative capabilities of LLMs, we utilize the above collected data to train the reward model in a supervised fine-tuning manner.
Specifically, for generative reward modeling, the reward model takes the response $y$ as input and directly predicts the corresponding reward score $r$. For critique-based reward modeling, the model is trained to first generate the critique $c$ and then predict the reward score $r$. The training objective is defined by the cross-entropy loss:
\begin{equation}
\mathcal{L} = -\frac{1}{N} \sum_{i=1}^{N} 
\begin{cases} 
\log P(r^{(i)} \mid y^{(i)}; \theta), & \text{For generative reward modeling,} \\
\log P(c^{(i)},r^{(i)} \mid y^{(i)}; \theta), & \text{For critique-based reward modeling,}
\end{cases}
\end{equation}
where $N$ is the number of training samples.

\subsection{Dynamic System Switch}

In existing research~\cite{kahneman2011thinking,jiang2024technical,yu2024distill}, search-based decoding approaches that improve response quality by scaling inference time are referred to as the System 2 thinking mode, whereas direct generation methods are termed the System 1 thinking mode~\cite{kahneman2011thinking}. While System 2 provides superior response quality, its substantially higher computational and temporal costs make it impractical for universal application, as not all user queries necessitate complex reasoning.
To balance efficiency and response quality, we propose a dynamic system switch mechanism that adaptively selects the appropriate thinking mode. The mechanism operates on two hierarchical levels: \emph{instance-level switch}, which determines the generation mode for the input question, and \emph{step-level switch}, which adjusts the generation mode for individual reasoning steps. We refer to our MCTS approach with a system switch mechanism as MCTSwitch.

\textbullet~\textbf{Instance-level switch.} This switch determines the thinking system for a given question based on the complexity evaluation of the question. We argue that simple questions can be directly handled with System 1 to ensure efficiency, while complex questions leverage System 2 to enhance quality.

\textbullet~\textbf{Step-level switch.} If the question is determined to use System 2 thinking mode, we further employ the step-level switch to evaluate whether the next reasoning step requires System 1 or 2 to achieve a trade-off between the efficiency and effectiveness based on the complexity and uncertainty of the current context.

\paratitle{Switch Model Training.} To achieve reliable system switch, we train a switch model by collecting a set of synthetic data similar to the reward model training.
For the training data of step-level switch, we assign labels (0 or 1) to each node in the search tree based on its value. Specifically, we define a threshold $\gamma$: if a node's value exceeds $\gamma$, we label its state as 1, which denotes requiring System 2 thinking mode as a large value indicates a higher likelihood of generating hallucinated text. Conversely, we label a node with value below $\gamma$ as 0, indicating that System 2 thinking mode is not required. Through this process, we obtain the thinking system labels for each step and utilize them as the training data for step-level switch.
For instance-level data, we use the policy model to directly generate responses to the given question. The questions with correct responses are labeled as 0 (not requiring System 2 thinking mode), while those with incorrect responses are labeled as 1 (requiring System 2 thinking mode). These labeled questions serve as the training data for instance-level switch.
The system switch model is then trained on the mixed instance-level and step-level training data using supervised fine-tuning, following the same training objective as the reward model.

\paratitle{Switch Model Inference.}
During inference, the system switch model first predicts the thinking system at the instance level. If the prediction is 0, the policy model adopts System 1 thinking mode to directly generate a response. If the prediction is 1, the policy model employs MCTS to perform deliberate reasoning for generating high-quality responses. At the expansion step of MCTS, the switch model performs step-level switch by assessing the state of the selected node and predicting the thinking system (\ie 0 or 1). For nodes not requiring System 2 thinking mode, a single sentence is generated as the child node; while for other nodes, the policy model follows the expansion process (Section~\ref{sec: MCTS}) to generate multiple child nodes. The threshold $\gamma$ controls the proportion of System 1 and 2 thinking modes, balancing efficiency and reasoning quality. The entire process of our proposed HaluSearch approach is presented in Algorithm~\ref{alg:switchmcts}.

\section{Experiments}

In this part, we detail the experimental setup and then highlight the main takeaways of our results.

\subsection{Experimental Setup}

\paratitle{Datasets and Metrics.}
We evaluate HaluSearch across multiple question-answering datasets in both English and Chinese. For English, we select HaluEval-QA~\cite{li2023halueval}, TruthfulQA~\cite{truthfulqa}, and SimpleQA~\cite{wei2024measuring}. For Chinese, we select HalluQA~\cite{halluqa}, ChineseSimpleQA~\cite{he2024chinese}, and ChineseFactEval~\cite{factool}. We use \emph{accuracy} as the evaluation metric. Specifically, we employ GPT-4 to assess the correctness of each model-generated response by comparing it with the corresponding ground truth for each question.

\paratitle{Baselines.}
Since HaluSearch explicitly performs a slow thinking generation process through MCTS during inference, we select the following inference-stage hallucination mitigation methods as baselines for comparison. In addition, we report the accuracy of direct generation by the policy models as the lower bounds for reference.

\textbullet~\textbf{Chain-of-Thought}~\cite{wei2022chain} prompts the model to generate intermediate reasoning steps before arriving at the final answer. In this work, we employ zero-shot CoT, which appends the phrase ``Let's think step by step.'' into the prompt.

\textbullet~\textbf{Self-Consistency}~\cite{wangself} samples multiple responses during inference and selects the most consistent response as the final answer.

\textbullet~\textbf{Best-of-N}~\cite{lightmanlet} is similar to self-consistency, which selects the best response through a reward model.

\textbullet~\textbf{Self-Refine}~\cite{madaan2024self} generates an initial response, evaluates it through feedback, and iteratively refines the response based on this feedback until a satisfactory version is achieved.

\textbullet~\textbf{ITI}~\cite{li2024inference} operates by shifting model activations during inference to enhance the truthfulness of the generated responses.

\paratitle{Implementation Details.}
We evaluate our approach and the compared baselines using Llama3.1-8B-Instruct~\cite{llama3.1} and Qwen2-7B-Instruct~\cite{qwen2} as policy models, with GPT-4 serving as the reward model. In the MCTS process, we set the number of nodes expanded per step to 10, perform 5 rollouts for each node, and limit the maximum number of simulations to 20. In the UCT algorithm, the weight $w$ to control the exploration and exploitation is set to 0.4. For Self-Consistency and Best-of-N, we sample 20 responses for each question. For ITI, we follow the original setting~\cite{li2024inference} by using models adjusted on the TruthfulQA dataset and report performance only on English datasets to ensure fairness. For all methods, the decoding temperature of the policy model is set to 0.9, with a 0-shot prompting configuration.

\subsection{Main Results}
The evaluation results of our method and the baselines are presented in Table~\ref{tab: main-exp}.

Firstly, prompt-based generation methods demonstrate improved response accuracy compared to direct generation. However, the extent of this improvement is inconsistent and varies across tasks. For instance, on the TruthfulQA dataset, Chain-of-Thought prompting achieves an accuracy of 33.50\% and Self-Refine attains 30.50\% for Llama3.1-8B-Instruct, representing improvements of 9.00\% and 6.00\% compared to direct generation, respectively. However, on the HaluEval-QA dataset, these methods show less pronounced improvements, with CoT achieving 36.80\% and Self-Refine performing even worse at 28.20\%. This limitation arises from the inherent capabilities of the policy models and their sensitivity to the provided prompts, preventing less capable models from engaging in genuine deliberation and effective self-refinement.

Secondly, inference-time intervention methods exhibit limited generalization when adjusting activations on specific datasets. For ITI, the models we use are designed to probe and adjust activations on the TruthfulQA dataset, achieving effective improvements on this dataset (\eg 37.50\% on the TruthfulQA dataset with Llama3.1-8B-Instruct as the policy model). However, this effectiveness diminishes on other datasets, often underperforming compared to direct generation. For instance, the accuracy of ITI drops to 2.00\% on the SimpleQA dataset, demonstrating its limited transferability in scenarios with scarce data.

Thirdly, strategies that generate multiple responses and leverage reward signals for selection can robustly enhance the quality of model outputs and effectively reduce hallucinations. Among all baselines, Self-Consistency and Best-of-N exhibit relatively strong performance across most datasets (\eg 36.00\% accuracy on ChineseFactEval for SC and 39.00\% on HaluEval-QA for BoN).
In comparison, HaluSearch achieves the best performance across all six Chinese and English datasets. By leveraging MCTS to model the response generation process as step-by-step reasoning, our method provides fine-grained reward signals for each generation step, enabling effective guidance and a balance between exploration and exploitation. This structured reasoning framework reduces error accumulation throughout the generation process, ultimately mitigating hallucinations in the final response, demonstrating the effectiveness of slow thinking and deliberate reasoning during inference.

\begin{table*}[t]
\caption{Evaluation results of \textbf{Llama3.1-8B-Instruct} and \textbf{Qwen2-7B-Instruct} on six English and Chinese datasets. \textbf{Bold} denotes the best results and \underline{underline} denotes the second best results.}
\centering
\small
\renewcommand{\arraystretch}{1.25}
\resizebox{\textwidth}{!}{
\begin{tabular}{l c c c c c c}
\toprule
\textbf{Methods} & 
\textbf{HaluEval-QA} & 
\textbf{TruthfulQA} & \textbf{SimpleQA} & \textbf{HalluQA} & \textbf{ChineseSimpleQA} & \textbf{ChineseFactEval} \\ 
\midrule
\rowcolor[gray]{0.9} \multicolumn{7}{c}{\textbf{Llama3.1-8B-Instruct}} \\
Direct Generation & 35.60 & 24.50 & 3.00 & 8.25 & 15.50 & 29.60 \\
CoT & 36.80 & 33.50 & 4.00 & 6.80 & 19.50 &32.00\\
SC & 35.00 & 39.00 & \underline{6.00} & 8.25 & 21.00 &\underline{36.00}\\
BoN & \underline{37.80} & \underline{43.50} & 5.50 & \underline{12.62} & \underline{29.00}&35.60\\
Self-Refine & 28.20  & 30.50 & 5.00 & 8.25 & 17.50 &33.60\\
ITI & 35.20 & 37.50 & 2.00 & - & - & -\\
MCTS & \textbf{45.40} & \textbf{47.50} & \textbf{8.50} & \textbf{16.50} & \textbf{30.50} & \textbf{40.80}\\
\midrule
\rowcolor[gray]{0.9} \multicolumn{7}{c}{\textbf{Qwen2-7B-Instruct}} \\
Direct Generation & 34.00 & 26.50 & 5.50 & 32.04 & 28.50 & 56.00 \\
CoT & 35.60 & \underline{38.00} & 4.50 & 33.50 & 29.00 & 54.40 \\
SC & 37.20 & 36.00 & 4.00 & 32.52 & 30.50& 59.20\\
BoN & \underline{39.00} & 37.50 & \underline{6.00} & \underline{35.44} & \underline{35.00} & \underline{64.00}\\
Self-Refine & 35.80 & 26.00 & \underline{6.00} & 26.21 & 30.50 & 48.80\\
ITI & 32.20  & 17.00 & 3.00 & - & - & -\\
MCTS & \textbf{43.69} & \textbf{45.07}& \textbf{12.50} & \textbf{43.69} & \textbf{36.00} & \textbf{70.40}\\
\bottomrule
\end{tabular}}
\label{tab: main-exp}
\end{table*}

\subsection{Reward Model Analysis}
Beyond employing GPT-4 as the reward model, we conduct experiments to evaluate the performance of our trained self-evaluation reward models. Specifically, we investigate the two reward modeling approaches described in Section~\ref{sec-reward-model}: 
\begin{enumerate}[label=(\arabic*)]
    \item \emph{Generative RM},  which is trained on the synthetic generative reward data and directly generates numerical scores to responses during evaluation; 
    \item \emph{Generative RM + Critic}, which first criticizes and analyzes the responses, and then generates a score based on the content of this feedback. 
\end{enumerate}
We sample 1,000 examples from the HaluEval-QA dataset and 500 examples from the TruthfulQA dataset to generate training data, reserving the remaining data for evaluation. After filtering, we obtain 52K samples for reward data and 38K samples for critique data. These datasets are used to train reward models based on Llama3.1-8B-Instruct and Qwen2-7B-Instruct. We evaluate their performance with different reward modeling approaches on the three English datasets, and the results are presented in Table~\ref{tab: selfrm}.

Compared to the upper bound established by GPT-4 RM, Generative RM achieves competitive performance after training, outperforming other baselines shown in Table~\ref{tab: main-exp}. Building on this, the Generative RM + Critic approach, which is trained using GPT-4-generated critiques and scores, shows substantial improvement over Generative RM. By introducing a critique step before scoring, this method achieves significant gains, particularly on the TruthfulQA dataset (50.00\% for Llama3.1-8B-Instruct and 50.50\% for Qwen2-7B-Instruct), even surpassing the GPT-4 RM when it only provides scores. These results indicate that incorporating a critique step can improve the scoring accuracy of Generative RM, thereby enabling more effective self-evaluation.

\begin{table}[tb]
\caption{Evaluation results of different self-evaluation reward modeling approaches on Llama3.1-8B-Instruct and Qwen2-7B-Instruct on HaluEval-QA dataset.}
\centering
\renewcommand{\arraystretch}{1.25}
\resizebox{0.6\textwidth}{!}{
\begin{tabular}{l c c c}
\toprule
\textbf{Reward Model} & 
\textbf{HaluEval-QA} & 
\textbf{TruthfulQA} & \textbf{SimpleQA} \\ 
\midrule
\rowcolor[gray]{0.9} \multicolumn{4}{c}{\textbf{Llama3.1-8B-Instruct}} \\
GPT-4 RM& 45.40 & 47.50 &  8.50 \\
\midrule
Generative RM  & 42.60 & 45.50 &  7.50 \\
Generative RM + Critic & 46.20 & 50.00 & 7.85 \\
\midrule
\rowcolor[gray]{0.9} \multicolumn{4}{c}{\textbf{Qwen2-7B-Isntruct}} \\
GPT-4 RM & 43.69 & 45.07 &  12.50\\
\midrule
Generative RM  & 40.40 & 40.00 & 6.50 \\
Generative RM + Critic& 42.80 & 50.50 & 8.50\\
\bottomrule
\end{tabular}
}
\label{tab: selfrm}
\end{table}

\begin{figure}[t]
	\centering
	\includegraphics[width=0.8\textwidth]{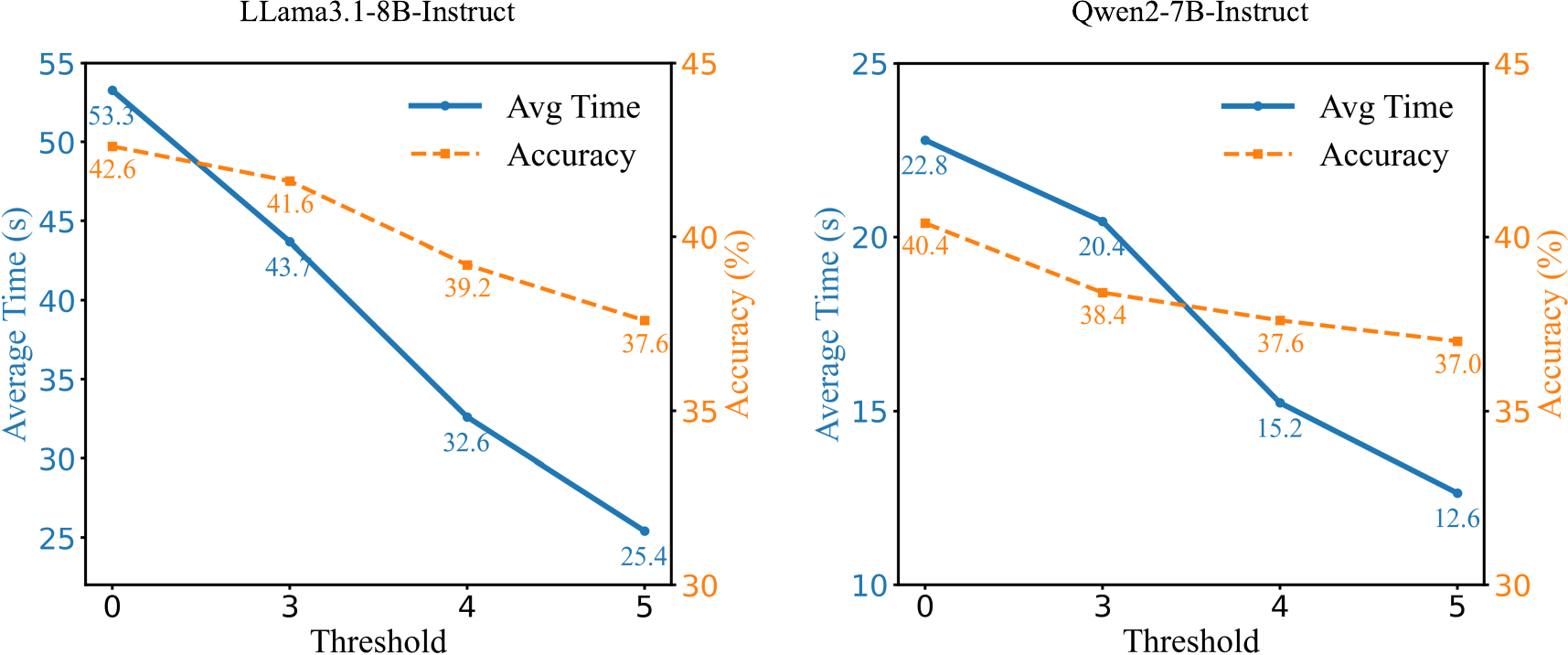}
	\caption{Impact of switch thresholds on time saving and accuracy on HaluEval-QA dataset.}
	\label{fig: switch}
\end{figure}

\subsection{System Switch Analysis}
To validate the effectiveness of the proposed dynamic system switch mechanism, we investigate the impact of slow thinking threshold $\gamma$ (\ie the proportion of responses generated in slow thinking mode) on hallucination rates and response efficiency. Specifically, we collect 10K training data from HaluEval-QA and TruthfulQA, categorize them based on different score thresholds (\eg $\gamma=3,4,5$), and label samples exceeding these thresholds for the slow thinking mode, as higher scores signify a higher likelihood of hallucination. We use the categorized datasets to train switch models optimized for different thresholds and evaluate their performance on the HaluEval-QA dataset. Here, both the reward model and the switch model are trained using Llama3.1-8B-Instruct and Qwen2-7B-Instruct, which also serve as the policy models.

In Figure~\ref{fig: switch}, we present the accuracy of generated responses and the average solving time per question under different switch thresholds. As shown, when the switch threshold is set to 0, corresponding to a 100\% slow thinking ratio, the model achieves its highest accuracy of 42.6\% with an average solving time of 53.3 seconds. We consider this the upper bound for model accuracy. As the switch threshold increases, the proportion of slow thinking gradually decreases, with only states exhibiting very high hallucination levels triggering slow thinking. This results in a significant reduction in the average solving time per question, accompanied by a slight decrease in response accuracy. This trade-off between time efficiency and accuracy becomes more pronounced with higher switch thresholds. For instance, at a threshold of 5, where the slow thinking ratio is minimal, the average solving time per question drops to 25.4 seconds, while the accuracy remains at 37.6\%. These observations demonstrate that our proposed system switch mechanism can effectively balance computational efficiency and response accuracy, allowing users to tailor the trade-off to specific task requirements.

\subsection{Further Analysis}

\begin{figure}[t]
	\centering
	\includegraphics[width=0.7\textwidth]{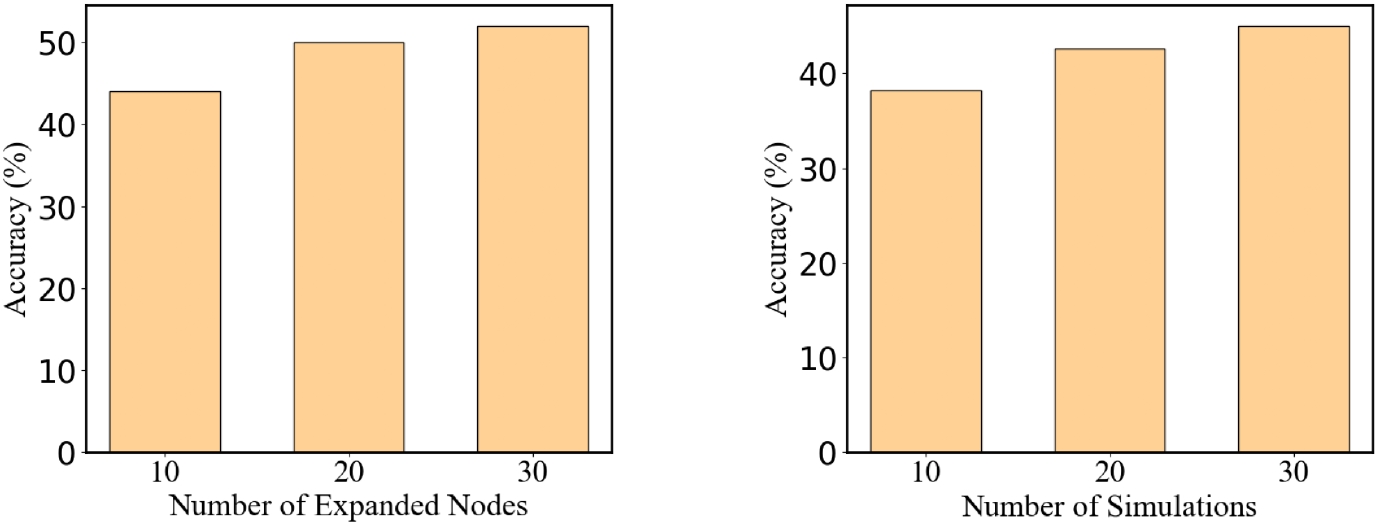}
	\caption{Results on HaluEval-QA w.r.t. the number of expanded nodes or the number of simulations.}
	\label{fig: parameter}
\end{figure}

\paratitle{Hyper-parameter Analysis.}
To validate the effectiveness of our proposed framework, we conduct a detailed analysis of its key hyper-parameters. Specifically, we analyze the influence of the number of expanded nodes per step and simulations on the performance of HaluSearch. Experiments are conducted on the HaluEval-QA dataset using Llama3.1-8B-Instruct as the policy model. We vary the number of expanded nodes in the set \{10, 20, 50\} while keeping the number of simulations fixed at 20, and vary the number of simulations in the set \{10, 20, 30\} while keeping the number of expanded nodes fixed at 10. The results are shown in Figure~\ref{fig: parameter}. As we can see, increasing the number of expanded nodes and simulations improves the performance of HaluSearch. This improvement is attributed to the expanded search space, which increases the likelihood of identifying the correct answer by sampling more potential responses. However, as the number of expanded nodes and simulations continues to grow, the performance gains decrease due to the inherent limitations of the internal knowledge of the policy model and the maximum scoring accuracy of the reward model.

\begin{figure*}[t]
	\centering
	\includegraphics[width=1.00\textwidth]{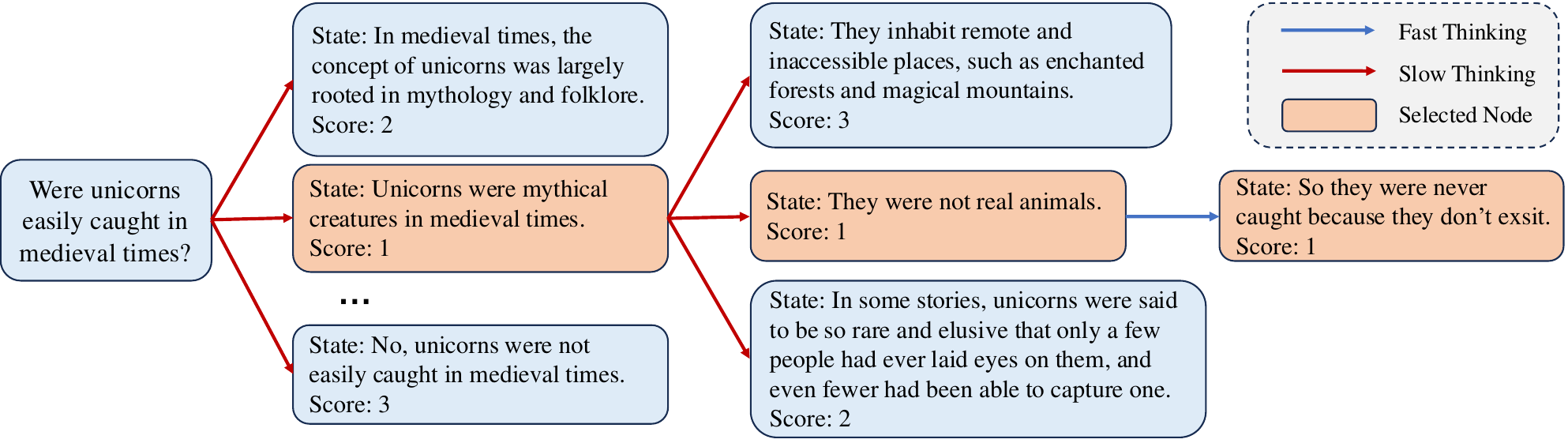}
	\caption{An example showing the deliberative reasoning process with system switching of HaluSearch in TruthfulQA.}
	\label{fig: casestudy}
        \vspace{-0.2cm}
\end{figure*}

\paratitle{Case Study.}
In Figure~\ref{fig: casestudy}, we present an example from the TruthfulQA dataset to illustrate the reasoning process of HaluSearch. Starting from the question as the root node (\ie \textit{``Were unicorns easily caught in medieval times?’’}), the policy model employs slow thinking mode to generate multiple reasoning steps. While the model accurately identifies that unicorns are mythical creatures, it also generates erroneous reasoning paths, such as treating unicorns as real animals that could potentially be caught.
The reward model evaluates these steps, assigning lower hallucination scores to accurate reasoning steps (\eg a score of 1 for the state: \textit{Unicorns were mythical creatures in medieval times’’}). Nodes with low hallucination scores are further expanded through slow thinking, yielding reliable intermediate reasoning steps (\eg \textit{They were not real animals’’}). Based on these intermediate states, fast thinking is applied to efficiently derive the final answer: \textit{``Unicorns don’t exist.’’}
This iterative process of reasoning enables the policy model to progressively reduce hallucinations, mitigate error accumulation, and achieve a balance between computational efficiency and response reliability.

\section{Conclusion}

In this work, we presented HaluSearch, a framework that integrates tree search-based algorithms (\eg MCTS) to enable explicit slow thinking processes in LLMs, aiming at mitigating hallucinations in their responses. The proposed approach models text generation as a step-by-step reasoning process, guided by a self-evaluation reward model that scores each step and selects the most reliable generation path.
To facilitate self-evaluation, we trained the reward model using data synthesized by the HaluSearch framework to assess the degree of hallucinations and provide reward signals.
Additionally, to improve efficiency, we introduced a dynamic system switch mechanism, which utilizes a trained switch model to enable LLMs to adaptively alternate between fast and slow thinking modes at both the instance and step levels.
Extensive experimental results demonstrated that our framework outperformed other inference-stage hallucination mitigation methods across a range of Chinese and English datasets.


\bibliographystyle{unsrt}
\bibliography{ref.bib}
\newpage

\section*{MCTSwitch Algorithm}
\label{app: mctswitch}

We formalize and present our proposed dynamic system switch mechanism (\ie MCTSwitch) in Algorithm~\ref{alg:switchmcts}.

\begin{algorithm}[th]
\caption{MCTSwitch}
\label{alg:switchmcts}
\begin{algorithmic}[1]
\STATE \textbf{Input:} policy model $\pi_\theta$, reward model $R$, switch model $\sigma_s$, number of expansions $K$, number of rollouts $m$, UCT weight $\omega$, max iterations $M$, reward threshold $r_{\text{th}}$, query $q$
\STATE \textbf{Initialize:} Root $s_0 \gets q$, $\mathcal{C}(s_0) = \emptyset$, $t \gets 0$
\STATE \textbf{Instance Switch:} $\text{instance\_mode} \gets \sigma_s(q)$
\IF{$\text{instance\_mode} = \text{slow}$} 
    \WHILE{$t < M$}
        \STATE $s_t \gets \text{Select}(\text{UCT}(s_0, \omega))$
        \STATE \textbf{Step Switch:} $\text{step\_mode} \gets \sigma_s(s_t)$
        \IF{$\text{step\_mode} = \text{slow}$}
            \STATE $\mathcal{C}(s_t) \gets \text{Expand}(s_t, \pi_\theta, K)$
            \FOR{$s_c \in \mathcal{C}(s_t)$}
                \STATE $\mathcal{C_\mathrm{r}}(s_c) \gets \text{Rollout}(s_c, \pi_\theta, m)$
                \STATE $r_{s_c} \gets \text{Avg}(\mathcal{C_\mathrm{r}}(s_c), R)$
                \IF{$r_{s_c} \geq r_{\text{th}}$ \text{and} $s_c$ \text{is terminal}}
                    \STATE \textbf{Break}
                \ENDIF
            \ENDFOR
        \ELSE
            \STATE $s_c \gets \pi_\theta(s_t)$
            \STATE $\mathcal{C_\mathrm{r}}(s_c) \gets \text{Rollout}(s_c, \pi_\theta, m)$
            \STATE $r_{s_c} \gets R(s_c)$
            \IF{$r_{s_c} \geq r_{\text{th}}$ \text{and} $s_c$ \text{is terminal}}
                \STATE \textbf{Break}
            \ENDIF
        \ENDIF
        \STATE \text{Backpropagate}$(s_0, \mathcal{C}(r_{s_c}))$
        \STATE $t \gets t + 1$
    \ENDWHILE
    \STATE $A \gets \text{BestNode}(s_0)$
\ELSE
    \STATE $A \gets \pi_\theta(q)$
\ENDIF
\STATE \textbf{Output:} $A$
\end{algorithmic}
\end{algorithm}

\end{document}